# Evidence with Uncertain Likelihoods

**Joseph Y. Halpern**
Cornell University
Ithaca, NY 14853 USA
halpern@cs.cornell.edu

**Riccardo Pucella**
Cornell University
Ithaca, NY 14853 USA
riccardo@cs.cornell.edu

## Abstract

An agent often has a number of hypotheses, and must choose among them based on observations, or outcomes of experiments. Each of these observations can be viewed as providing *evidence* for or against various hypotheses. All the attempts to formalize this intuition up to now have assumed that associated with each hypothesis $h$ there is a *likelihood function* $\mu_h$, which is a probability measure that intuitively describes how likely each observation is, conditional on $h$ being the correct hypothesis. We consider an extension of this framework where there is uncertainty as to which of a number of likelihood functions is appropriate, and discuss how one formal approach to defining evidence, which views evidence as a function from priors to posteriors, can be generalized to accommodate this uncertainty.

## 1 Introduction

An agent often has a number of hypotheses, and must choose among them based on observations, or outcomes of experiments. Each of these observations can be viewed as providing *evidence* for or against various hypotheses. The following simple example illustrates the situation.

**Example 1.1** Suppose that Alice and Bob each have a coin. Alice's coin is double-headed, Bob's coin is fair. Charlie knows all of this. Alice and Bob give their coin to some third party, Zoe, who chooses one of the coins, and tosses it. Charlie is not privy to Zoe's choice, but gets to see the outcome of the toss. Charlie is interested in two events (which are called hypotheses in this context):

  $A$: the coin is Alice's coin

  $B$: the coin is Bob's coin.

Now Charlie observes the coin land heads. What can he say about the probability of the events $A$ and $B$? If Charlie has no prior probability on $A$ and $B$, then he can draw no conclusions about their posterior probability; the probability of $A$ could be any number in $[0, 1]$. The same remains true if the coin lands heads 100 times in a row. □

Clearly Charlie learns something from seeing 100 (or even one) coin toss land heads. This has traditionally been modeled in terms of *evidence*: the more times Charlie sees heads, the more evidence he has for the coin being heads. There have been a number of ways of modeling evidence in the literature; see [Kyburg 1983] for an overview. All of them make use of the *likelihood function*. More precisely, they assume that for each hypothesis $h$ of interest, there is a probability $\mu_h$ (called a likelihood function) on the space of possible observations. In the example above, if the coin is tossed once, the two possible observations are $heads$ and $tails$. Clearly $\mu_A(heads) = 1/2$ and $\mu_B(heads) = 1$. If the coin is tossed 100 times, then there are $2^{100}$ possible observations (sequences of coin tosses). Again, $\mu_A$ and $\mu_B$ put obvious probabilities on this space. In particular, if $100heads$ is the observation of seeing 100 heads in a row, then $\mu_A(100heads) = 1/2^{100}$ and $\mu_B(100heads) = 1$. Most of the approaches compute the relative evidence of a particular observation $ob$ for two hypotheses $A$ and $B$ by comparing $\mu_A(ob)$ and $\mu_B(ob)$.

Our goal in this paper is to understand what can be done when an hypothesis $h$ does not determine a unique probability $\mu_h$. To understand the issues that arise, consider the following somewhat contrived variant of Example 1.1.

**Example 1.2** Suppose that Alice has two coins, one that is double-headed and one that is biased 3/4 towards heads, and chooses which one to give Zoe. Again, Zoe chooses either Alice's coin or Bob's coin and tosses it. Charlie, who knows the whole setup, sees the coin land heads. What does this tell him about the likelihood that the coin tossed was Alice's? □

The problem is that now we do not have a probability $\mu_A$ on observations corresponding to the coin being Alice's coin, since Charlie does not know if Alice's coin is double-headed or biased $3/4$ towards heads. It seems that there is an obvious solution to this problem. We simply split the hypothesis "the coin is Alice's coin" into two hypotheses:

$A_1$: the coin is Alice's coin and it is double-headed

$A_2$: the coin is Alice's coin and it is the biased coin.

Now we can certainly apply standard techniques for computing evidence to the three hypotheses $A_1$, $A_2$, and $B$. The question now is what do the answers tell us about the evidence in favor of the coin being Alice's coin?

Situations like that in Example 1.2 arise frequently. For example, consider a robot equipped with an unreliable sensor for navigation. This sensor returns the distance to the wall in front of the robot, with some known error. For simplicity, suppose that distances are measured in integral units $0, 1, 2, \ldots$, and that if the wall is at distance $m$, then the sensor will return a reading of $m - 1$ with probability $1/4$, a reading of $m$ with probability $1/2$, and a reading of $m+1$ with probability $1/4$. Suppose the robot wants to stop if it is exactly close to the wall, where "close" is interpreted as being within 3 units of the wall, and go forward if it is farther than 3 units. So again, we have two hypotheses of interest. However, while for each specific distance $m$ we have a probability $\mu_m$ on sensor readings, we do not have a probability on sensor readings corresponding to the hypothesis *far*: "the robot is farther than 3 from the wall". While standard techniques will certainly give us the weight of evidence of a particular sensor reading for the hypothesis "the robot is distance $m$ from the wall", it is not clear what the weight of evidence should be for the hypothesis *far*.

To examine the problem carefully, we consider one particular approach for determining the weight of evidence, due to Shafer [1982], which is a generalization of a method advocated by Good [1950]. Let an evidence space $\mathcal{E}$ consist of a set $\mathcal{H}$ of possible hypotheses, a set $\mathcal{O}$ of observations, and a probability $\mu_h$ on observations for each $h \in \mathcal{H}$. We take the weight of evidence for hypothesis $h$ provided by observation $ob$ in evidence space $\mathcal{E}$, denoted $w_\mathcal{E}(ob, h)$, to be

$$w_\mathcal{E}(ob, h) = \frac{\mu_h(ob)}{\sum_{h' \in \mathcal{H}} \mu_{h'}(ob)}.$$

It is easy to see that $w_\mathcal{E}(ob, \cdot)$ acts like a probability on $\mathcal{H}$, in that $\sum_{h \in \mathcal{H}} w_\mathcal{E}(ob, h) = 1$. With this definition, it is easy to compute the weight of evidence for Alice's coin when Charlie sees heads in Example 1.1 is $2/3$, and the weight of evidence when Charlie sees 100 heads is $2^{100}/(2^{100} + 1)$. As expected, the more often Charlie sees heads, the more evidence he has in favor of the coin being double-headed (provided that he does not see tails).

In Example 1.2, if we consider the three hypotheses $A_1$, $A_2$, and $B$, then the weight of evidence for $A_1$ when Charlie sees heads is $1/(1 + 3/4 + 1/2) = 4/9$; similarly, the weight of evidence for $A_2$ is $1/3$ and the weight of evidence for $B$ is $2/9$. Since weight of evidence acts like a probability, it might then seem reasonable to take the weight of evidence for $A$ (the coin used was Alice's coin) to be $4/9 + 1/3 = 7/9$. (Indeed, this approach was implicitly suggested in our earlier paper [Halpern and Pucella 2003a].) But is this reasonable? A first hint that it might not be is the observation that the weight of evidence for $A$ is higher in this case than it is in the case where Alice certainly had a double-headed coin.

To analyze this issue, we need an independent way of understanding what evidence is telling us. As observed by Halpern and Fagin [1992], weight of evidence can be viewed as a function from priors to posteriors. That is, given a prior on hypotheses, we can combine the prior with the weight of evidence to get the posterior. In particular, if there are two hypotheses, say $H_1$ and $H_2$, the weight of evidence for $H_1$ is $\alpha$, and the prior probability of $H_1$ is $\beta$, then the posterior probability of $H_1$ (that is, the probability of $H_1$ in light of the evidence) is

$$\frac{\alpha\beta}{\alpha\beta + (1-\alpha)(1-\beta)}.$$

Thus, for example, by deciding to perform an action when the weight of evidence for $A$ is $2/3$ (i.e., after Charlie has seen the coin land heads once), Charlie is assured that, if the prior probability of $A$ is at least $.01$, then the posterior probability of $A$ is at least $2/11$; similarly, after Charlie has seen 100 heads, if the prior probability of $A$ is at least $.01$, then the posterior probability of $A$ is at least $2^{100}/(2^{100} + 99)$.

But now consider the situation in Example 1.2. Again, suppose that the prior probability of $A$ is at least $.01$. Can we conclude that the posterior probability of $A$ is at least $.01(7/9)/(.01(7/9) + .99(2/9)) = 7/205$? As we show, we cannot. The calculation $(\alpha\beta)/(\alpha\beta + (1-\alpha)(1-\beta))$ is appropriate only when there are two hypotheses. If the hypotheses $A_1$ and $A_2$ have priors $\alpha_1$ and $\alpha_2$ and weights of evidence $\beta_1$ and $\beta_2$, then the posterior probability of $A$ is

$$\frac{\alpha_1\beta_1 + \alpha_2\beta_2}{\alpha_1\beta_1 + \alpha_2\beta_2 + (1-\alpha_1-\alpha_2)(1-\beta_1-\beta_2)},$$

which is in general quite different from

$$\frac{(\alpha_1+\alpha_2)(\beta_1+\beta_2)}{(\alpha_1+\alpha_2)(\beta_1+\beta_2) + (1-\alpha_1-\alpha_2)(1-\beta_1-\beta_2)}.$$

Moreover, it is easy to show that if $\beta_1 > \beta_2$ (as is the case here), then the posterior of $A$ is somewhere in the interval

$$\left[\frac{\alpha_2\beta_2}{\alpha_2\beta_2 + (1-\alpha_2)(1-\beta_2)}, \frac{\alpha_1\beta_1}{\alpha_1\beta_1 + (1-\alpha_1)(1-\beta_1)}\right].$$

That is, we get a lower bound on the posterior by acting as if the only possible hypotheses are $A_2$ and $B$, and we get an upper bound by acting as if the only possible hypotheses are $A_1$ and $B$.

In this paper, we generalize this observation by providing a general approach to dealing with weight of evidence when the likelihood function is unknown. In the special case when the likelihood function is known, our approach reduces to Shafer's approach. Roughly speaking, the idea is to consider all possible evidence spaces consistent with the information. The intuition is that one of them is the right one, but the agent trying to ascribe a weight of evidence does not know which. For example, in Example 1.2, the evidence space either involves hypotheses $\{A_1, B\}$ or hypotheses $\{A_2, B\}$: either Alice's first coin is used or Alice's second coin is used. We can then compute the weight of evidence for Alice's coin being used with respect to each evidence space. This gives us a range of possible weights of evidence, which can be used for decision making in a way that seems most appropriate for the problem at hand (by considering the max, the min, or some other function of the range).

The advantage of this approach is that it allows us to consider cases where there are correlations between the likelihood functions. For example, suppose that, in the robot example, the robot's sensor was manufactured at one of two factories. The sensors at factory 1 are more reliable than those of factory 2. Since the same sensor is used for all readings, the appropriate evidence space either uses all likelihood functions corresponding to factory 1 sensors, or all likelihood functions corresponding to factory 2 sensors.

The rest of this paper is organized as follows. In Section 2, we review Shafer's approach to dealing with evidence. In Section 3, we show how to extend it so as to deal with situation where the likelihood function is uncertain, and argue that our approach is reasonable. In Section 4, we consider how to combine evidence in this setting. We conclude in Section 5. The proofs of our technical results are deferred to the full paper.

## 2 Evidence: A Review

We briefly review the notion of evidence and its formalization by Shafer [1982], using some terminology from [Halpern and Pucella 2003b].

We start with a set $\mathcal{H}$ of hypotheses, which we take to be mutually exclusive and exhaustive; thus, exactly one hypothesis holds at any given time. We also have a set $\mathcal{O}$ of *observations*, which can be understood as outcomes of experiments that can be made. Finally, we assume that for each hypothesis $h \in \mathcal{H}$, there is a probability $\mu_h$ (often called a *likelihood function*) on the observations in $\mathcal{O}$. This is formalized as an *evidence space* $\mathcal{E} = (\mathcal{H}, \mathcal{O}, \boldsymbol{\mu})$, where $\mathcal{H}$ and $\mathcal{O}$ are as above, and $\boldsymbol{\mu}$ is a *likelihood mapping*, which assigns to every hypothesis $h \in \mathcal{H}$ a probability measure $\boldsymbol{\mu}(h) = \mu_h$. (For simplicity, we often write $\mu_h$ for $\boldsymbol{\mu}(h)$, when the former is clear from context.)

For an evidence space $\mathcal{E}$, the weight of evidence for hypothesis $h \in \mathcal{H}$ provided by observation $ob$, written $w_\mathcal{E}(ob, h)$, is

$$w_\mathcal{E}(ob, h) = \frac{\mu_h(ob)}{\sum_{h' \in \mathcal{H}} \mu_{h'}(ob)}. \quad (1)$$

The weight of evidence $w_\mathcal{E}$ is not defined by (1) for an observation $ob$ such that $\sum_{h \in \mathcal{H}} \mu_h(ob) = 0$. Intuitively, this means that the observation $ob$ is impossible. In the literature on evidence it is typically assumed that this case never arises. More precisely, it is assumed that all observations are possible, so that for every observation $ob$, there is an hypothesis $h$ such that $\mu_h(ob) > 0$. For simplicity, we make the same assumption here. (We remark that in some application domains this assumption holds because of the structure of the domain, without needing to be assumed explicitly; see [Halpern and Pucella 2003b] for an example.)

The measure $w_\mathcal{E}$ always lies between 0 and 1, with 1 indicating that the observation provides full evidence for the hypothesis. Moreover, for each fixed observation $ob$ for which $\sum_{h \in \mathcal{H}} \mu_h(ob) > 0$, $\sum_{h \in \mathcal{H}} w_\mathcal{E}(ob, h) = 1$, and thus the weight of evidence $w_\mathcal{E}$ looks like a probability measure for each $ob$. While this has some useful technical consequences, one should not interpret $w_\mathcal{E}$ as a probability measure. It is simply a way to assign a weight to hypotheses given observations, and, as we shall soon see, can be seen as a way to update a prior probability on the hypotheses into a posterior probability on those hypotheses, based on the observations made.

**Example 2.1** In Example 1.1, the set $\mathcal{H}$ of hypotheses is $\{A, B\}$; the set $\mathcal{O}$ of observations is simply $\{heads, tails\}$, the possible outcomes of a coin toss. From the discussion following the description of the example, it follows that $\boldsymbol{\mu}$ assigns the following likelihood functions to the hypotheses: since $\mu_A(heads)$ is the probability that the coin landed heads if it is Alice's coin (i.e., if it is double-headed), then $\mu_A(heads) = 1$ and $\mu_A(tails) = 0$. Similarly, $\mu_B(heads)$ is the probability that the coin lands heads if it is fair, so $\mu_B(heads) = 1/2$ and $\mu_B(tails) = 1/2$. This can be summarized by the following table:

| $\boldsymbol{\mu}$ | $A$ | $B$ |
|---|---|---|
| $heads$ | 1 | 1/2 |
| $tails$ | 0 | 1/2 |

Let

$$\mathcal{E} = (\{A, B\}, \{heads, tails\}, \boldsymbol{\mu}).$$

A straightforward computation shows that $w_\mathcal{E}(heads, A) = 2/3$ and $w_\mathcal{E}(heads, B) = 1/3$. Intuitively, the coin landing heads provides more evidence for the hypothesis $A$ than the hypothesis $B$. Similarly, $w(tails, A) = 0$ and $w(tails, A) = 1$. Thus, the coin landing tail indicates that the coin must be fair. This information can be represented by the following table:

| $w_{\mathcal{E}}$ | $A$ | $B$ |
|---|---|---|
| heads | 2/3 | 1/3 |
| tails | 0 | 1 |

□

It is possible to interpret the weight function $w$ as a prescription for how to update a prior probability on the hypotheses into a posterior probability on those hypotheses, after having considered the observations made [Halpern and Fagin 1992]. There is a precise sense in which $w_{\mathcal{E}}$ can be viewed as a function that maps a prior probability $\mu_0$ on the hypotheses $\mathcal{H}$ to a posterior probability $\mu_{ob}$ based on observing $ob$, by applying Dempster's Rule of Combination [Shafer 1976]. That is,

$$\mu_{ob} = \mu_0 \oplus w_{\mathcal{E}}(ob, \cdot), \qquad (2)$$

where $\oplus$ combines two probability distributions on $\mathcal{H}$ to get a new probability distribution on $\mathcal{H}$ as follows:

$$(\mu_1 \oplus \mu_2)(H) = \frac{\sum_{h \in H} \mu_1(h)\mu_2(h)}{\sum_{h \in \mathcal{H}} \mu_1(h)\mu_2(h)}. \qquad (3)$$

(Strictly speaking, $\oplus$ is defined for set functions, that is, functions with domain $2^{\mathcal{H}}$. We have defined $w_{\mathcal{E}}(ob, \cdot)$ as a function with domain $\mathcal{H}$, but is is clear from (3) that this is all that is really necessary to compute $\mu_0 \oplus w_{\mathcal{E}}(ob, \cdot)$ in our case.)

Bayes' Rule is the standard way of updating a prior probability based on an observation, but it is only applicable when we have a joint probability distribution on both the hypotheses and the observations, something which we did not assume we had. Dempster's Rule of Combination essentially "simulates" the effects of Bayes's rule. The relationship between Dempster's Rule and Bayes' Rule is made precise by the following well-known theorem.

**Proposition 2.2** [Halpern and Fagin 1992] *Let $\mathcal{E} = (\mathcal{H}, \mathcal{O}, \mu)$ be an evidence space. Suppose that $P$ is a probability on $\mathcal{H} \times \mathcal{O}$ such that $P(\mathcal{H} \times \{ob\}|\{h\} \times \mathcal{O}) = \mu_h(ob)$ for all $h \in \mathcal{H}$ and all $ob \in \mathcal{O}$. Let $\mu_0$ be the probability on $\mathcal{H}$ induced by marginalizing $P$; that is, $\mu_0(h) = P(\{h\} \times \mathcal{O})$. For $ob \in \mathcal{O}$, let $\mu_{ob} = \mu_0 \oplus w_{\mathcal{E}}(ob, \cdot)$. Then $\mu_{ob}(h) = P(\{h\} \times \mathcal{O}|\mathcal{H} \times \{ob\})$.*

In other words, when we do have a joint probability on the hypotheses and observations, then Dempster's Rule of Combination gives us the same result as a straightforward application of Bayes' Rule.

## 3 Evidence with Uncertain Likelihoods

In Example 1.1, each of the two hypotheses $A$ and $B$ determines a likelihood function. However, in Example 1.2, the hypothesis $A$ does not determine a likelihood function. By viewing it as the compound hypothesis $\{A_1, A_2\}$, as we did in the introduction, we can construct an evidence space with a set $\{A_1, A_2, B\}$ of hypotheses. We then get the following likelihood mapping $\boldsymbol{\mu}$:

| $\boldsymbol{\mu}$ | $A_1$ | $A_2$ | $B$ |
|---|---|---|---|
| heads | 1 | 3/4 | 1/2 |
| tails | 0 | 1/4 | 1/2 |

Taking

$$\mathcal{E} = (\{A_1, A_2, B\}, \{heads, tails\}, \boldsymbol{\mu}),$$

we can compute the following weights of evidence:

| $w_{\mathcal{E}}$ | $A_1$ | $A_2$ | $B$ |
|---|---|---|---|
| heads | 4/9 | 1/3 | 2/9 |
| tails | 0 | 1/3 | 2/3 |

If we are now given prior probabilities for $A_1$, $A_2$, and $B$, we can easily use Proposition 2.2 to compute posterior probabilities for each of these events, and then add the posterior probabilities of $A_1$ and $A_2$ to get a posterior probability for $A$.

But what if we are given only a prior probability $\mu_0$ for $A$ and $B$, and are not given probabilities for $A_1$ and $A_2$? As observed in the introduction, if we define $w_{\mathcal{E}}(heads, A) = w_{\mathcal{E}}(heads, A_1) + w_{\mathcal{E}}(heads, A_2) = 7/9$, and then try to compute the posterior probability of $A$ given that heads is observed by naively applying the equation in Proposition 2.2, that is, taking by $\mu_{heads}(A) = (\mu_0 \oplus w_{\mathcal{E}}(heads, \cdot))(A)$, we get an inappropriate answer. In particular, the answer is not the posterior probability in general.

To make this concrete, suppose that $\mu_0(A) = .01$. Then, as observed in the introduction, a naive application of this equation suggests that the posterior probability of $A$ is 7/205. But suppose that in fact $\mu_0(A_1) = \alpha$ for some $\alpha \in [0, .01]$. Then applying Proposition 2.2, we see that $\mu_{heads}(A_1) = \alpha(4/9)/(\alpha(4/9) + (.01 - \alpha)(1/3) + .99(2/9)) = 4\alpha/(\alpha + 2.01)$. It is easy to check that $4\alpha/(\alpha + 2.01) = 7/205$ iff $\alpha = 1407/81300$. That is, the naive application of the equation in Proposition 2.2 is correct only if we assume a particular (not terribly reasonable) value for the prior probability of $A_1$.

We now present one approach to dealing with the problem, and argue that it is reasonable.

Define a *generalized evidence space* to be a tuple $\mathcal{G} = (\mathcal{H}, \mathcal{O}, \Delta)$, where $\Delta$ is a set of likelihood mappings. Note for future reference that we can associate with the generalized evidence space $\mathcal{G} = (\mathcal{H}, \mathcal{O}, \Delta)$ the set $\mathcal{S}(\mathcal{G}) =$

$\{(\mathcal{H}, \mathcal{O}, \boldsymbol{\mu}) \mid \boldsymbol{\mu} \in \Delta\}$ of evidence spaces. Thus, given a generalized evidence space $\mathcal{G}$, we can define the *generalized weight of evidence* $w_\mathcal{G}$ to be the set $\{w_\mathcal{E} : \mathcal{E} \in \mathcal{S}(\mathcal{G})\}$ of weights of evidence. We often treat $w_\mathcal{G}$ as a set-valued function, writing $w_\mathcal{G}(ob, h)$ for $\{w(ob, h) \mid w \in w_\mathcal{G}\}$.

Just as we can combine a prior with the weight of evidence to get a posterior in a standard evidence spaces, given a generalized evidence space, we can combine a prior with a generalized weight of evidence to get a set of posteriors. Given a prior probability $\mu_0$ on a set $\mathcal{H}$ of hypotheses and a generalized weight of evidence $w_\mathcal{G}$, let $\mathcal{P}_{\mu_0, ob}$ be the set of posterior probabilities on $\mathcal{H}$ corresponding to an observation $ob$ and prior $\mu_0$, computed according to Proposition 2.2:

$$\mathcal{P}_{\mu_0, ob} = \{\mu_0 \oplus w(ob, \cdot) \mid w \in w_\mathcal{G}\}. \qquad (4)$$

**Example 3.1** The generalized evidence space for Example 1.2, where Alice's coin is unknown, is

$$\mathcal{G} = (\{A, B\}, \{heads, tails\}, \{\boldsymbol{\mu}_1, \boldsymbol{\mu}_2\}),$$

where $\boldsymbol{\mu}_1(A) = \mu_{A_1}$, $\boldsymbol{\mu}_2(A) = \mu_{A_2}$, and $\boldsymbol{\mu}_1(B) = \boldsymbol{\mu}_2(B) = \mu_B$. Thus, the first likelihood mapping corresponds to Alice's coin being double-headed, and the second corresponds to Alice's coin being biased $3/4$ towards heads. Then $w_\mathcal{G} = \{w_1, w_2\}$, where $w_1(heads, A) = 2/3$ and $w_2(heads, A) = 3/5$. Thus, if $\mu_0(A) = \alpha$, then $\mathcal{P}_{\mu_0, heads}(A) = \{\frac{3\alpha}{\alpha+2}, \frac{2\alpha}{\alpha+1}\}$. □

We have now given two approaches for capturing the situation in Example 1.2. The first involves refining the set of hypotheses —that is, replacing the hypothesis $A$ by $A_1$ and $A_2$—and using a standard evidence space. The second involves using a generalized evidence space. How do they compare?

To make this precise, we need to first define what a refinement is. We say that the evidence space $(\mathcal{H}', \mathcal{O}, \boldsymbol{\mu}')$ *refines*, or *is a refinement of*, the generalized evidence space $(\mathcal{H}, \mathcal{O}, \Delta)$ if there exists a surjection $g : \mathcal{H}' \to \mathcal{H}$ such that $\boldsymbol{\mu} \in \Delta$ if and only if, for all $h \in \mathcal{H}$, there exists some $h' \in g^{-1}(h)$ such that $\boldsymbol{\mu}(h) = \boldsymbol{\mu}'(h')$. That is, taking $\mathcal{P}_h = \{\boldsymbol{\mu}'(h') \mid h' \in g^{-1}(h)\}$, we must have $\Delta = \prod_{h \in \mathcal{H}} \mathcal{P}_h$. Intuitively, the hypothesis $h \in \mathcal{H}$ is refined to the set of hypothesis $g^{-1}(h) \subseteq \mathcal{H}'$; moreover, each likelihood function $\boldsymbol{\mu}(h)$ in a likelihood mapping $\boldsymbol{\mu} \in \Delta$ is the likelihood function $\boldsymbol{\mu}'(h')$ for some hypothesis $h'$ refining $h$. For example, the evidence space $\mathcal{E}$ at the beginning of this section (corresponding to Example 1.2) is a refinement of the generalized evidence space $\mathcal{G}$ in Example 3.1; the required surjection $g : \{A_1, A_2, B\} \to \{A, B\}$ maps $A_1$ and $A_2$ to $A$, and $B$ to $B$. A prior $\mu_0'$ on $\mathcal{H}'$ extends a prior $\mu_0$ on $\mathcal{H}$ if for all $h$,

$$\mu_0'(g^{-1}(h)) = \mu_0(h).$$

Let $Ext(\mu_0)$ consist of all priors on $\mathcal{H}'$ that extend $\mu_0$. Recall that given a set $\mathcal{P}$ of probability measures, the *lower probability* $\mathcal{P}_*(U)$ of a set $U$ is $\inf\{\mu(U) \mid \mu \in \mathcal{P}\}$ and its *upper probability* $\mathcal{P}^*(U)$ is $\sup\{\mu(U) \mid \mu \in \mathcal{P}\}$ [Halpern 2003].

**Proposition 3.2** *Let $\mathcal{G} = (\mathcal{H}, \mathcal{O}, \Delta)$ be a generalized evidence space and let $\mathcal{E} = (\mathcal{H}', \mathcal{O}, \boldsymbol{\mu})$ be a refinement of $\mathcal{G}$. For all $ob \in \mathcal{O}$ and all $h \in \mathcal{H}$, we have*

$$(\mathcal{P}_{\mu_0, ob})^*(h) = \\ \{\mu_0' \oplus w_\mathcal{E}(ob, \cdot) \mid \mu_0' \in Ext(\mu_0)\}^*(g^{-1}(h))$$

*and*

$$(\mathcal{P}_{\mu_0, ob})_*(h) = \\ \{\mu_0' \oplus w_\mathcal{E}(ob, \cdot) \mid \mu_0' \in Ext(\mu_0)\}_*(g^{-1}(h)).$$

In other words, if we consider the sets of posteriors obtained by either (1) updating a prior probability $\mu_0$ by the generalized weight of evidence of an observation in $\mathcal{G}$ or (2) updating the set of priors extending $\mu_0$ by the weight of evidence of the same observation in $\mathcal{E}$, the bounds on those two sets are the same. Therefore, this proposition shows that, given a generalized evidence space $\mathcal{G}$, if there an evidence space $\mathcal{E}$ that refines it, then the weight of evidence $w_\mathcal{G}$ gives us essentially the same information as $w_\mathcal{E}$. But is there always an evidence space $\mathcal{E}$ that refines a generalized evidence space? That is, can we always understand a generalized weight of evidence in terms of a refinement? As we now show, we cannot always do this.

Let $\mathcal{G}$ be a generalized evidence space $(\mathcal{H}, \mathcal{O}, \Delta)$. Note that if $\mathcal{E}$ refines $\mathcal{G}$ then, roughly speaking, the likelihood mappings in $\Delta$ consist of all possible ways of combining the likelihood functions corresponding to the hypotheses in $\mathcal{H}$. We now formalize this property. A set $\Delta$ of likelihood mappings is *uncorrelated* if there exist sets of probability measures $\mathcal{P}_h$ for each $h \in \mathcal{H}$ such that

$$\Delta = \prod_{h \in \mathcal{H}} \mathcal{P}_h = \{\boldsymbol{\mu} \mid \boldsymbol{\mu}(h) \in \mathcal{P}_h \text{ for all } h \in \mathcal{H}\}.$$

(We say $\Delta$ is *correlated* if it is not uncorrelated.) A generalized evidence space $(\mathcal{H}, \mathcal{O}, \Delta)$ is uncorrelated if $\Delta$ is uncorrelated.

Observe that if $(\mathcal{H}', \mathcal{O}, \boldsymbol{\mu}')$ refines $(\mathcal{H}, \mathcal{O}, \Delta)$, then $(\mathcal{H}, \mathcal{O}, \Delta)$ is uncorrelated since, as observed above, $\Delta = \prod_{h \in \mathcal{H}} \mathcal{P}_h$, where $\mathcal{P}_h = \{\boldsymbol{\mu}'(h') \mid h' \in g^{-1}(h)\}$. Not only is every refinement uncorrelated, but every uncorrelated evidence space can be viewed as a refinement.

**Proposition 3.3** *Let $\mathcal{G}$ be a generalized evidence space. There exists an evidence space $\mathcal{E}$ that refines $\mathcal{G}$ if and only if $\mathcal{G}$ is uncorrelated.*

Thus, if a situation can be modeled using an uncorrelated generalized evidence space, then it can also be modeled by refining the set of hypotheses and using a simple evidence space. The uncorrelated case has a further advantage. It leads to simple formula for calculating the posterior in the special case that there are only two hypotheses (which is the case that has been considered most often in the literature, often to the exclusion of other cases).

Given a generalized evidence space $\mathcal{G} = (\mathcal{H}, \mathcal{O}, \Delta)$ and the corresponding generalized weight of evidence $w_\mathcal{G}$, we can define *upper* and *lower* weights of evidence, determined by the maximum and minimum values in the range, somewhat analogous to the notions of upper and lower probability. Define the *upper weight of evidence function* $\overline{w}_\mathcal{G}$ by taking

$$\overline{w}_\mathcal{G}(ob, h) = \sup\{w(ob, h) \mid w \in w_\mathcal{G}\}.$$

Similarly, define the *lower weight of evidence function* $\underline{w}_\mathcal{G}$ by taking

$$\underline{w}_\mathcal{G}(ob, h) = \inf\{w(ob, h) \mid w \in w_\mathcal{G}\}.$$

These upper and lower weights of evidence can be used to compute the bounds on the posteriors obtained by updating a prior probability via the generalized weight of evidence of an observation, in the case where $\mathcal{G}$ is uncorrelated, and when there are two hypotheses.

**Proposition 3.4** *Let $\mathcal{G} = (\mathcal{H}, \mathcal{O}, \Delta)$ be an uncorrelated generalized evidence space.*

(a) *The following inequalities hold:*

$$(\mathcal{P}_{\mu_0, ob})^*(h) \leq$$
$$\frac{\overline{w}_\mathcal{G}(ob, h)\mu_0(h)}{\overline{w}_\mathcal{G}(ob, h)\mu_0(h) + \sum_{h' \neq h} \underline{w}_\mathcal{G}(ob, h')\mu_0(h')}; \quad (5)$$

$$(\mathcal{P}_{\mu_0, ob})_*(h) \geq$$
$$\frac{\underline{w}_\mathcal{G}(ob, h)\mu_0(h)}{\underline{w}_\mathcal{G}(ob, h)\mu_0(h) + \sum_{h' \neq h} \overline{w}_\mathcal{G}(ob, h')\mu_0(h')}. \quad (6)$$

*If $|\mathcal{H}| = 2$, we get equalities in (5) and (6).*

(b) *The following equalities hold:*

$$\overline{w}_\mathcal{G}(ob, h) = \frac{(\mathcal{P}_h)^*(ob)}{(\mathcal{P}_h)^*(ob) + \sum_{h' \neq h} (\mathcal{P}_{h'})_*(ob)};$$

$$\underline{w}_\mathcal{G}(ob, h) = \frac{(\mathcal{P}_h)_*(ob)}{(\mathcal{P}_h)_*(ob) + \sum_{h' \neq h} (\mathcal{P}_{h'})^*(ob)},$$

*where $\mathcal{P}_h = \{\boldsymbol{\mu}(h) \mid \boldsymbol{\mu} \in \Delta\}$, for all $h \in \mathcal{H}$.*

Thus, if have an uncorrelated generalized evidence space with two hypotheses, we can compute the bounds on the posteriors $\mathcal{P}_{\mu_0, ob}$ in terms of upper and lower weights of evidence using Proposition 3.4(a), which consists of equalities in that case. Moreover, we can compute the upper and lower weights of evidence using Proposition 3.4(b). As we now show, the inequalities in Proposition 3.4(a) can be strict if there are more than two hypotheses.

**Example 3.5** Let $\mathcal{H} = \{D, E, F\}$ and $\mathcal{O} = \{X, Y\}$, and consider the two probability measures $\mu_1$ and $\mu_2$, where $\mu_1(X) = 1/3$ and $\mu_2(X) = 2/3$. Let $\mathcal{G} = (\mathcal{H}, \mathcal{O}, \Delta)$, where $\Delta = \{\boldsymbol{\mu} \mid \boldsymbol{\mu}(h) \in \{\mu_1, \mu_2\}\}$. Clearly, $\Delta$ is uncorrelated. Let $\mu_0$ be the uniform prior on $\mathcal{H}$, so that $\mu_0(D) = \mu_0(E) = \mu_0(F) = 1/3$. Using Proposition 3.4(b), we can compute that the upper and lower weights of evidence are as given in the following tables:

| $\overline{w}_\mathcal{E}$ | D | E | F |
|---|---|---|---|
| X | 1/2 | 1/2 | 1/2 |
| Y | 1/2 | 1/2 | 1/2 |

| $\underline{w}_\mathcal{E}$ | D | E | F |
|---|---|---|---|
| X | 1/5 | 1/5 | 1/5 |
| Y | 1/5 | 1/5 | 1/5 |

The uniform measure is the identity for $\oplus$, and therefore $\mu_0 \oplus w(ob, \cdot) = w(ob, \cdot)$. It follows that $\mathcal{P}_{\mu_0, X} = \{w(X, \cdot) \mid w \in w_\mathcal{G}\}$. Hence, $(\mathcal{P}_{\mu_0, X})^*(D) = 1/2$ and $(\mathcal{P}_{\mu_0, X})_*(D) = 1/5$. But the right-hand sides of (5) and (6) are $5/9$ and $1/6$, respectively, and similarly for hypotheses $E$ and $F$. Thus, in this case, the inequalities in Proposition 3.4(a) are strict. □

While uncorrelated generalized evidence spaces are certainly of interest, correlated spaces arise in natural settings. To see this, first consider the following somewhat contrived example.

**Example 3.6** Consider the following variant of Example 1.2. Alice has two coins, one that is double-headed and one that is biased 3/4 towards heads, and chooses which one to give Zoe. Bob also has two coins, one that is fair and one that is biased 2/3 towards tails, and chooses which one to give Zoe. Zoe chooses one of the two coins she was given and tosses it. The hypotheses are $\{A, B\}$ and the observations are $\{heads, tails\}$, as in Example 1.2. The likelihood function $\mu_1$ for Alice's double-headed coin is given by $\mu_1(heads) = 1$, while the likelihood function $\mu_2$ for Alice's biased coin is given by $\mu_2(heads) = 3/4$. Similarly, the likelihood function $\mu_3$ for Bob's fair coin is given by $\mu_3(heads) = 1/2$, and the likelihood function $\mu_4$ for Bob's biased coin is given by $\mu_4(heads) = 1/3$.

If Alice and Bob each make their choice of which coin to give Zoe independently, we can use the following generalized evidence space to model the situation:

$$\mathcal{G}_1 = (\{A, B\}, \{heads, tails\}, \Delta_1),$$

where

$$\Delta_1 = \{(\mu_1, \mu_3), (\mu_1, \mu_4), (\mu_2, \mu_3), (\mu_2, \mu_4)\}.$$

Clearly, $\Delta_1$ is uncorrelated, since it is equal to $\{\mu_1, \mu_2\} \times \{\mu_3, \mu_4\}$.

On the other hand, suppose that Alice and Bob agree beforehand that either Alice gives Zoe her double-headed coin and Bob gives Zoe his fair coin, or Alice gives Zoe her biased coin and Bob gives Zoe his biased coin. This situation can be modeled using the following generalized evidence space:

$$\mathcal{G}_2 = (\{A, B\}, \{heads, tails\}, \Delta_2),$$

where

$$\Delta_2 = \{(\mu_1, \mu_3), (\mu_2, \mu_4)\}.$$

Here, note that $\Delta_2$ is a correlated set of likelihood mappings. □

While this example is artificial, the example in the introduction, where the robot's sensors could have come from either factory 1 or factory 2, is a perhaps more realistic case where correlated evidence spaces arise. The key point here is that these examples show that we need to go beyond just refining hypotheses to capture a situation.

## 4 Combining Evidence

An important property of Shafer's [1982] representation of evidence is that it is possible to combine the weight of evidence of independent observations to obtain the weight of evidence of a sequence of observations. The purpose of this section is to show that our framework enjoys a similar property, but, rather unsurprisingly, new subtleties arise due to the presence of uncertainty. For simplicity, in this section we concentrate exclusively on combining the evidence of a sequence of two observations; the general case follows in a straightforward way.

Recall how combining evidence is handled in Shafer's approach. Let $\mathcal{E} = (\mathcal{H}, \mathcal{O}, \boldsymbol{\mu})$ be an evidence space. We define the likelihood functions $\mu_h$ on pairs of observations, by taking $\mu_h(\langle ob_1, ob_2 \rangle) = \mu_h(ob_1)\mu_h(ob_2)$. In other words, the probability of observing a particular sequence of observations given $h$ is the product of the probability of making each observation in the sequence. Thus, we are implicitly assuming that the observations are independent. It is well known (see, for example, [Halpern and Fagin 1992, Theorem 4.3]) that Dempster's Rule of Combination can be used to combine evidence; that is,

$$w_\mathcal{E}(\langle ob_1, ob_2 \rangle, \cdot) = w_\mathcal{E}(ob_1, \cdot) \oplus w_\mathcal{E}(ob_2, \cdot).$$

If we let $\mu_0$ be a prior probability on the hypotheses, and $\mu_{\langle ob_1, ob_2 \rangle}$ be the probability on the hypotheses after observing $ob_1$ and $ob_2$, we can verify that

$$\mu_{\langle ob_1, ob_2 \rangle} = \mu_0 \oplus w_\mathcal{E}(\langle ob_1, ob_2 \rangle, \cdot).$$

Here we are assuming that exactly one hypothesis holds, and it holds each time we make an observation. That is, if Zoe picks the double-headed coin, she uses it for both coin tosses.

**Example 4.1** Recall Example 2.1, where Alice just has a double-headed coin and Bob just has a fair coin. Suppose that Zoe, after being given the coins and choosing one of them, tosses it twice, and it lands heads both times. It is straightforward to compute that

| $w_\mathcal{E}$ | $A$ | $B$ |
|---|---|---|
| $\langle heads, heads \rangle$ | 4/5 | 1/5 |
| $\langle heads, tails \rangle$ | 0 | 1 |
| $\langle tails, heads \rangle$ | 0 | 1 |
| $\langle tails, tails \rangle$ | 0 | 1 |

Not surprisingly, if either of the observations is *tails*, the coin cannot be Alice's. In the case where the observations are $\langle heads, heads \rangle$, the evidence for the coin being Alice's (that is, double-headed) is greater than if a single heads is observed, since from Example 2.1, $w_\mathcal{E}(heads, A) = 2/3$. This agrees with our intuition that seeing two heads in a row provides more evidence for a coin to be double-headed than if a single heads is observed. □

How should we combine evidence for a sequence of observations when we have a generalized evidence space? That depends on how we interpret the assumption that the "same" hypothesis holds for each observation. In a generalized evidence space, we have possibly many likelihood functions for each hypothesis. The real issue is whether we use the same likelihood function each time we evaluate an observation, or whether we can use a different likelihood function associated with that hypothesis. The following examples show that this distinction can be critical.

**Example 4.2** Consider Example 1.2 again, where Alice has two coins (one double-headed, one biased toward heads), and Bob has a fair coin. Alice chooses a coin and gives it to Zoe; Bob gives his coin to Zoe. As we observed, there are two likelihood functions in this case, which we called $w_1$ and $w_2$; $w_1$ corresponds to Alice's coin being double-headed, and $w_2$ corresponds to the coin being biased 3/4 towards heads. Suppose that Zoe tosses the coin twice. Since she is tossing the same coin, it seems most appropriate to consider the generalized weight of evidence

$$\{w' \mid w'(\langle ob_1, ob_2 \rangle, \cdot) = w_i(ob_1, \cdot) \oplus w_i(ob_2, \cdot),$$
$$i \in \{1, 2\}\}.$$

On the other hand, suppose Zoe first chooses whether she will always use Alice's or Bob's coin. If she chooses Bob, then she obviously uses his coin for both tosses. If she chooses Alice, before each toss, she asks Alice for a coin and tosses it; however, she does not have to use the same coin of Alice's for each toss. Now the likelihood function associated with each observation can change. Thus, the appropriate generalized weight of evidence is

$$\{w' \mid w'(\langle ob_1, ob_2\rangle, \cdot) = w_i(ob_1, \cdot) \oplus w_j(ob_2, \cdot),$$
$$i, j \in \{1, 2\}\}.$$

□

Fundamentally, combining evidence in generalized evidence spaces relies on Dempster's rule of combination, just like in Shafer's approach. However, as Example 4.2 shows, the exact details depends on our understanding of the experiment. While the first approach used in Example 4.2 seems more appropriate in most cases that we could think of, we suspect that there will be cases where something like the second approach may be appropriate.

## 5 Conclusion

In the literature on evidence, it is generally assumed that there is a single likelihood function associated with each hypothesis. There are natural examples, however, which violate this assumption. While it may appear that a simple step of refining the set of hypotheses allows us to use standard techniques, we have shown that this approach can lead to counterintuitive results when evidence is used as a basis for making decisions. To solve this problem, we proposed a generalization of a popular approach to representing evidence. This generalization behaves correctly under updating, and gives the same bounds on the posterior probability as that obtained by refining the set of hypotheses when there is no correlation between the various likelihood functions for the hypotheses. As we show, this is the one situation where we can identify a generalized evidence space with the space obtained by refining the hypotheses. One advantage of our approach is that we can also reason about situations where the likelihood functions are correlated, something that cannot be done by refining the set of hypotheses.

We have also looked at how to combine evidence in a generalized evidence space. While the basic ideas from standard evidence spaces carry over, that is, the combination is essentially obtained using Dempster's rule of combination, the exact details of how this combination should be performed depend on the specifics of how the likelihood functions change for each observation. A more detailed dynamic model would be helpful in understanding the combination of evidence in a generalized evidence space setting; we leave this exploration for future work.

**Acknowledgments**

Work supported in part by NSF under grants CTC-0208535 and ITR-0325453, by ONR under grant N00014-02-1-0455, by the DoD Multidisciplinary University Research Initiative (MURI) program administered by the ONR under grants N00014-01-1-0795 and N00014-04-1-0725, and by AFOSR under grant F49620-02-1-0101. The second author was also supported in part by AFOSR grants F49620-00-1-0198 and F49620-03-1-0156, National Science Foundation Grants 9703470 and 0430161, and ONR Grant N00014-01-1-0968.